\documentclass[journal]{IEEEtran}

\ifCLASSINFOpdf
\else
\usepackage[dvips]{graphicx}
\fi
\usepackage{url}

\usepackage{amsfonts}

\usepackage{amsmath}
\usepackage{graphicx}
\usepackage{epstopdf}
\usepackage{tabularx}
\usepackage{multirow}
\usepackage{textcomp}

\usepackage{lipsum} 
\usepackage{enumitem}
\setlist{nosep}

\graphicspath{{./image/}}

\usepackage{color}

\usepackage{bm}
\usepackage{calc}
\usepackage{balance}
\usepackage{makecell}

\usepackage{enumitem}
\usepackage{url}
\usepackage{cite}

\usepackage{xcolor,colortbl}

\usepackage{environ}

\NewEnviron{NORMAL}{%
	\scalebox{2}{$\BODY$}
}

\NewEnviron{SMALL}{%
	\scalebox{0.75}{$\BODY$}
}

\NewEnviron{HUGE}{%
	\scalebox{5}{$\BODY$}
}

\usepackage{sidecap, caption}

\usepackage[final]{microtype}

\begin{document}
	
	\title{Improving GANs for Speech Enhancement}
	
	\author{Huy Phan$^\ast$, Ian V. McLoughlin, Lam Pham, Oliver Y. Ch\'en, Philipp Koch, Maarten De Vos, Alfred Mertins
		\thanks{HP is with Queen Mary University of London, UK. IVM is with Singapore Institute of Technology, Singapore. LP is with the University of Kent, UK. OYC is with the University of Oxford, UK. MDV is with KU Leuven, Belgium. PK and AM are with the University of L\"ubeck, Germany.}
		\thanks{This research received funding from the Flemish Government (AI Research Program). Maarten De Vos is affiliated to Leuven.AI - KU Leuven institute for AI, B-3000, Leuven, Belgium.}
		\thanks{$^\ast$Correspondance email: \texttt{h.phan@qmul.ac.uk}}}
	
	\markboth{This letter has been accepted for publication in IEEE Signal Processing Letters}
	{This letter has been accepted for publication in IEEE Signal Processing Letters}
	\maketitle
	\begin{abstract}
		Generative adversarial networks (GAN) have recently
		been shown to be efficient for speech enhancement. However, most,
		if not all, existing speech enhancement GANs (SEGAN) make
		use of a single generator to perform one-stage enhancement
		mapping. In this work, we propose to use multiple generators that are chained to perform multi-stage enhancement mapping, which gradually refines the noisy input signals in a stage-wise fashion. Furthermore, we study two scenarios: (1) the generators share their parameters and (2) the generators' parameters are independent. The former constrains the generators to learn a common mapping that is iteratively applied at all enhancement stages and results in a small model footprint. On the contrary, the latter allows the generators to flexibly learn different enhancement mappings at different stages of the network at the cost of an increased model size. We demonstrate that the proposed multi-stage enhancement approach outperforms the one-stage SEGAN baseline, where the independent generators lead to more favorable results than the tied generators. The source code is available at http://github.com/pquochuy/idsegan.
	\end{abstract}
	
	\vspace{-0.05cm}

	\IEEEpeerreviewmaketitle

	\vspace{-0.35cm}
	\section{Introduction}
	\label{sec:intro}
	\vspace{-0.1cm}
	
	The goal of speech enhancement is to improve the quality and intelligibility of speech which are degraded by background noise \cite{Loizou2013, Yang2005}. Speech enhancement can serve as a front-end to improve performance of an automatic speech recognition system \cite{Weninger2015}. It also plays an important role in applications like communication systems, hearing aids, and cochlear implants in which contaminated speech needs to be enhanced prior to signal amplification to reduce discomfort \cite{Yang2005}. Significant progress on this research topic has been made with the involvement of deep learning paradigms. Deep neural networks (DNNs) \cite{Xu2015, Kumar2016}, convolutional neural networks (CNNs) \cite{Park2017, Mamun2019}, and recurrent neural networks (RNNs) \cite{Weninger2015, Erdogan2015} have been exploited either to produce the enhanced signal directly via a regression form \cite{Xu2015, Park2017} or to estimate the contaminating noise, which is subtracted from the noisy signal to obtain the enhanced signal \cite{Mamun2019}. Significant improvements on speech enhancement performance have been reported by these deep-learning based methods over more conventional ones, such as Wiener filtering \cite{Lim1978}, spectral subtraction \cite{Boll1979} or minimum mean square error (MMSE) estimation \cite{Ephraim1985,Gerkmann2011}.
	
	There exists a class of generative methods relying on GANs \cite{Goodfellow2014}, which have been demonstrated to be efficient for speech enhancement \cite{Pascual2017, Pascual2019, Higuchi2017, Qin2018, Li2018, Donahue2018}.
	When GANs are used for this task, the enhancement mapping is accomplished by the generator $G$ whereas the discriminator $D$, by discriminating between real and fake signals, transmits information to $G$ so that $G$ can learn to produce output that resembles the realistic distribution of the clean signals.  Using GANs, speech enhancement has been done using either magnitude spectrum input \cite{Li2018} or raw waveform input \cite{Pascual2017, Pascual2019}.
	
	\begin{figure} [!t]
		\centering
		\includegraphics[width=.75\linewidth]{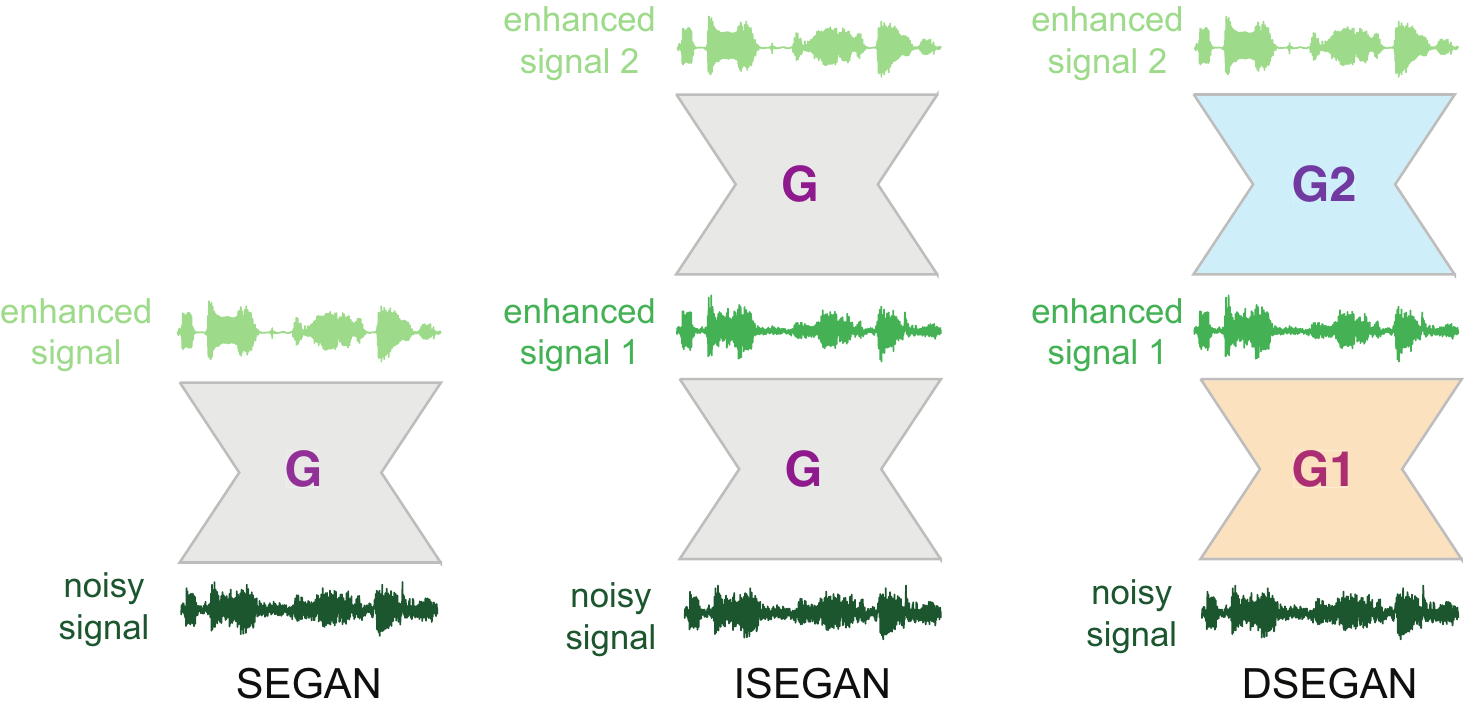}
		\vspace{-0.15cm}
		\caption{Illustration of SEGAN with a single generator $G$, ISEGAN ($N\!=\!2$) with the shared generators $G$, and DSEGAN ($N=2$) with two independent generators $G1$ and $G2$.}
		\label{fig:segan}
		\vspace{-0.35cm}
	\end{figure}
	
	Existing speech enhancement GAN (SEGAN) systems share a common feature -- the enhancement mapping is accomplished via a single stage by a single generator $G$ \cite{Pascual2017, Pascual2019,Li2018}, which may not be optimal. Here, we aim to divide the enhancement process into multiple stages and accomplish it via multiple enhancement mappings, one at each stage. Each of the mappings is realized by a generator, and the generators are chained to enhance a noisy input signal gradually, step by step, to yield an enhanced signal. By doing so, a generator is tasked to refine or correct the output produced by its predecessor. We hypothesize that it would be better to carry out multi-stage enhancement mapping rather than a single-stage one as in prior works \cite{Pascual2017, Pascual2019,Li2018}. We then propose two new SEGAN frameworks, namely iterated SEGAN (ISEGAN) and deep SEGAN (DSEGAN) as illustrated in Fig. \ref{fig:segan}, to study two scenarios: (1) using a common mapping for all the enhancement stages and (2) using independent mappings at different enchancement stages. In the former the generators' parameters are tied and parameter sharing constrains ISEGAN's generators to learn a common mapping (i.e. the generators apply the same mapping iteratively). The latter's generators have independent parameters, allowing them to learn different enhacement mappings flexibly. Note that, due to parameter sharing, ISEGAN's footprint is expected to be smaller than that of DSEGAN.
	
	We will demonstrate that the proposed method obtains more favorable results than the SEGAN baseline \cite{Pascual2017} on both objective and subjective evaluation metrics and that learning independent mappings with DSEGAN leads to better performance than learning a common one with ISEGAN.
	
	\vspace{-0.25cm}
	\section{SEGAN}
	\label{sec:gan}
	\vspace{-0.1cm}
	
	Given a dataset $\mathcal{X}\!=\!\{\!(\mathbf{x}_1, \mathbf{\tilde{x}}_1), (\mathbf{x}_2, \mathbf{\tilde{x}}_2), \ldots, (\mathbf{x}_N, \mathbf{\tilde{x}}_N)\!\}$ consisting of $N$ pairs of raw signals: clean speech signal $\mathbf{x}$ and noisy speech signal $\mathbf{\tilde{x}}$, speech enhancement is to find a mapping $f(\mathbf{\tilde{x}})\!:\!\mathbf{\tilde{x}}\!\mapsto\!\mathbf{x}$ to map the noisy signal $\mathbf{\tilde{x}}$ to the clean signal $\mathbf{x}$. Conforming to GAN's principle \cite{Goodfellow2014}, SEGAN proposed in \cite{Pascual2017} has its generator $G$ tasked for the enhancement mapping. Presented with the noisy signal $\mathbf{\tilde{x}}$ together with the latent representation $\mathbf{z}$, $G$ produces the enhanced signal $\mathbf{\hat{x}} = G(\mathbf{z}, \mathbf{\tilde{x}})$. The discriminator $D$ of SEGAN receives a pair of signals as input. $D$ learns to classify the pair $(\mathbf{x}, \mathbf{\tilde{x}})$ as real and the pair $(\mathbf{\hat{x}}, \mathbf{\tilde{x}})$ as fake while $G$ tries to fool $D$ such that $D$ classifies the pair $(\mathbf{\hat{x}}, \mathbf{\tilde{x}})$ as real. The objective function of SEGAN reads
	\begin{align}
		\min_G\max_D V\!(D,G) = \mathbb{E}_{\mathbf{x},\mathbf{\tilde{x}}\sim p_{\text{data}}(\mathbf{x},\mathbf{\tilde{x}})}\!\log\!D(\mathbf{x},\mathbf{\tilde{x}}) \nonumber \\ +\,\mathbb{E}_{\mathbf{z}\sim p_{\mathbf{z}}(\mathbf{z}),\mathbf{\tilde{x}}\sim p_{\text{data}}(\mathbf{\tilde{x}})}\!\log(1 - D(G(\mathbf{z},\mathbf{\tilde{x}}), \mathbf{\tilde{x}})).
	\end{align}
	
	To improve the stability, SEGAN further employs least-squares GAN (LSGAN) \cite{Mao2017} to replace the discriminator $D$'s cross-entropy loss by the least-square loss. The least-squares objective functions of $D$ and $G$ are explicitly written as
	\begin{align}
		\min_DV_{\text{LS}}(D) = &\frac{1}{2}\mathbb{E}_{\mathbf{x},\mathbf{\tilde{x}}\sim p_{\text{data}}(\mathbf{x},\mathbf{\tilde{x}})}(D(\mathbf{x},\mathbf{\tilde{x}}) - 1)^2 \nonumber \\ &+ \frac{1}{2}\mathbb{E}_{\mathbf{z}\sim p_{\mathbf{z}}(\mathbf{z}),\mathbf{\tilde{x}}\sim p_{\text{data}}(\mathbf{\tilde{x}})}D(G(\mathbf{z},\mathbf{\tilde{x}}), \mathbf{\tilde{x}})^2,
		\label{eq:D_objective}
	\end{align}
	\vspace{-0.5cm}
	{\small
		\begin{align}
			\min_GV_{\text{LS}}(G)=&\frac{1}{2}\mathbb{E}_{\mathbf{z}\sim p_{\mathbf{z}}(\mathbf{z}),\mathbf{\tilde{x}}\sim p_{\text{data}}(\mathbf{\tilde{x}})}(D(G(\mathbf{z},\mathbf{\tilde{x}}), \mathbf{\tilde{x}})-1)^2 \nonumber\\ &+ \lambda|| G(\mathbf{z},\mathbf{\tilde{x}}) - \mathbf{x}||_1,
			\label{eq:G_objective}
		\end{align}
	}respectively. In (\ref{eq:G_objective}), $\ell_1$ distance between the clean sample $\mathbf{x}$ and the generated sample $G(\mathbf{z},\mathbf{\tilde{x}})$ is included to encourage the generator $G$ to generate more fine-grained and realistic results \cite{Pascual2017,Isola2017,Pathak2016}. The influence of the $\ell_1$-norm term is regulated by the hyper-parameter $\lambda$ which was set to $\lambda=100$ in \cite{Pascual2017}.
	
	\vspace{-0.1cm}
	\section{Iterated SEGAN and Deep SEGAN}
	\vspace{-0.05cm}
	Quan \emph{et al.} \cite{Quan2018} showed that using an additional generator chained to the generator of a GAN leads to better image-reconstruction performance. In light of this, instead of using the single-stage enhancement mapping with one generator as in SEGAN, 
	we propose to learn multiple mappings with a chain of $N$ generators $\mathfrak{G}\!=\!G_1\!\!\rightarrow\!\! G_2\!\!\rightarrow\!\!\ldots\!\!\rightarrow\!\!G_N$ with $N\!>\!1$ to perform multi-stage enhancement. We study both the cases when a common mapping is learned and shared by all the stages (i.e. ISEGAN) and when independent mappings are learned at different stages (i.e. DSEGAN).  In ISEGAN, the generators share their parameters (i.e. they are realized by a common generator $G$) and can be viewed as an iterated generator with the \emph{number of iterations} of $N$. In constrast, DSEGAN's generators are independent and can be viewed as a deep generator with the \emph{depth} of $N$. 
	ISEGAN and DSEGAN with $N\!=\!2$ are illustrated alongside SEGAN in Fig. \ref{fig:segan}. Both ISEGAN and DSEGAN reduce to SEGAN when $N\!=\!1$.
	
	\begin{figure} [!t]
		\centering
		\includegraphics[width=0.8\linewidth]{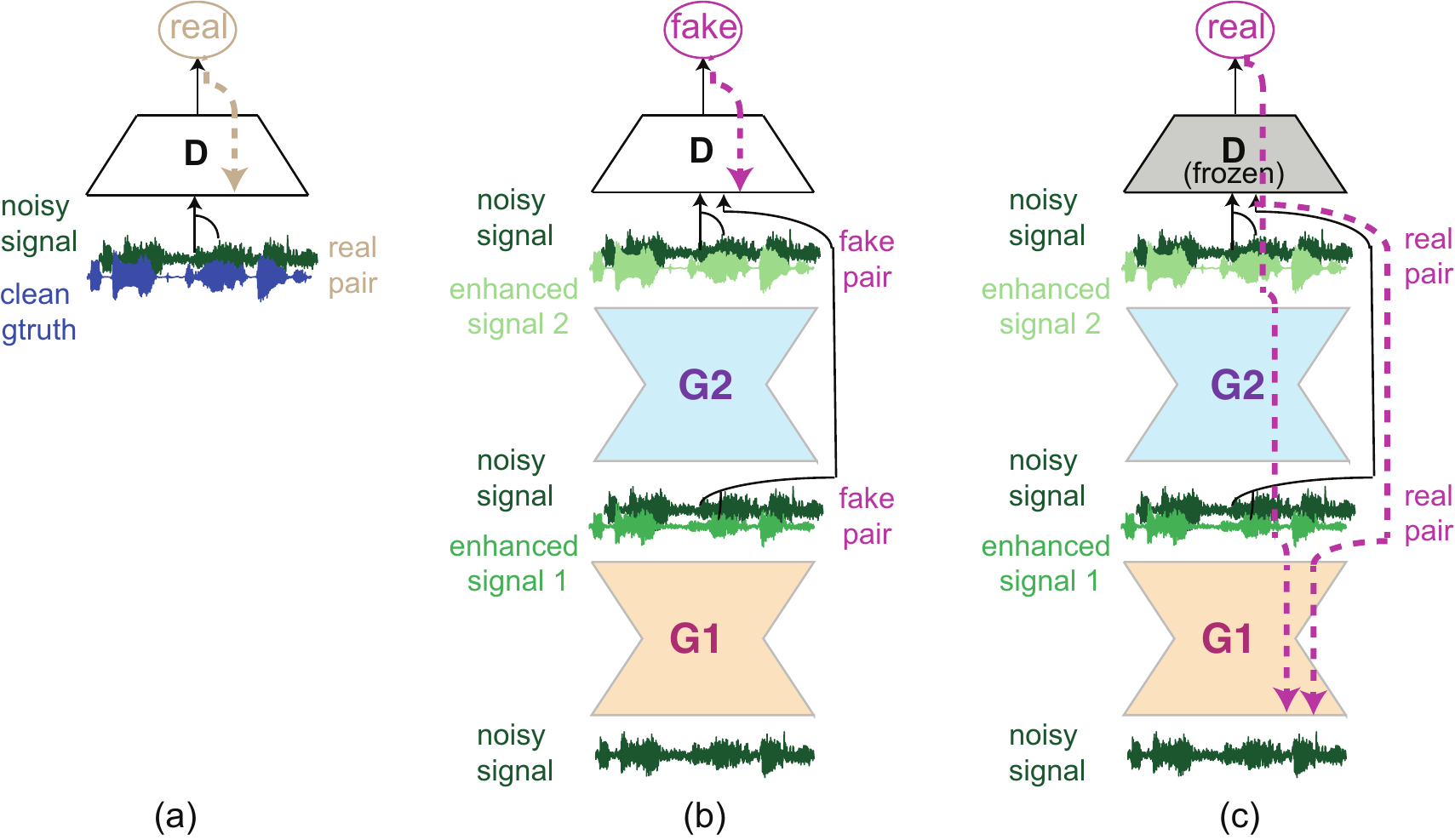}
		\vspace{-0.15cm}
		\caption{Adversarial training with two generators. The discriminator $D$ is learned to classify the pair $(\mathbf{x}, \mathbf{\tilde{x}})$ as real (a), and all the pairs $(\mathbf{\hat{x}}_1, \mathbf{\tilde{x}})$, $(\mathbf{\hat{x}}_2, \mathbf{\tilde{x}})$, $\ldots$, $(\mathbf{\hat{x}}_N, \mathbf{\tilde{x}})$ as fake (b). The generators $G_1$ and $G_2$ are learned to fool $D$ so that $D$ classifies the pairs $(\mathbf{\hat{x}}_1, \mathbf{\tilde{x}})$, $(\mathbf{\hat{x}}_2, \mathbf{\tilde{x}})$, $\ldots$, $(\mathbf{\hat{x}}_N, \mathbf{\tilde{x}})$ as real (c). Dashed lines represent the flow of gradient backdrop.}
		\label{fig:segan_learning}
		\vspace{-0.2cm}
	\end{figure}
	
	At the enhancement stage $n$, $1 \le n \le N$, the generator $G_n$ receives the output $\mathbf{\hat{x}}_{n-1}$ of its predecessor $G_{n-1}$ together with the latent representation $\mathbf{z}_{n}$ and is expected to produce a better enhanced signal $\mathbf{\hat{x}}_{n}$:
	\begin{align}
		\mathbf{\hat{x}}_n &=  G_n(\mathbf{z}_n, \mathbf{\hat{x}}_{n-1}), \mbox{~~~} 1 \le n \le N.
	\end{align}
	Note that $\mathbf{\hat{x}}_0\!\equiv\!\mathbf{\tilde{x}}$. The output of the last generator $G_N$ is considered as the final enhanced signal, i.e. $\mathbf{\hat{x}}\!\equiv\!\mathbf{\hat{x}}_N$, which is expected to be of better quality than all the intermediate enhanced versions. The outputs of the generators can be interpreted as different checkpoints and by forcing the 
	ground-truth between the checkpoints, we encourage the chained generators to produce gradually better enhancement results. 
	
	To enforce the generators in the chain $\mathfrak{G}$ to learn a proper mapping for signal enhancement, the discriminator $D$ is tasked  to classify the pair $(\mathbf{x}, \mathbf{\tilde{x}})$ as real while all $N$ pairs $(\mathbf{\hat{x}}_1, \mathbf{\tilde{x}})$, $(\mathbf{\hat{x}}_2, \mathbf{\tilde{x}})$, $\ldots$, $(\mathbf{\hat{x}}_N, \mathbf{\tilde{x}})$ as fake, as illustrated in Fig.~\ref{fig:segan_learning} for the case of $N~=~2$. The least-squares objective functions of $D$ and $\mathfrak{G}$ are given as
	{\small \begin{align}
			& \min_DV_{\text{LS}}(D) = \,\frac{1}{2}\mathbb{E}_{\mathbf{x},\mathbf{\tilde{x}}\sim p_{\text{data}}(\mathbf{x},\mathbf{\tilde{x}})}(D(\mathbf{x},\mathbf{\tilde{x}}) - 1)^2 \nonumber \\ &+\!\sum\nolimits_{n=1}^N\!\frac{1}{2N}\mathbb{E}_{\mathbf{z}_n\sim p_{\mathbf{z}}(\mathbf{z}),\mathbf{\tilde{x}}\sim p_{\text{data}}(\mathbf{\tilde{x}})}\!D(G_n(\mathbf{z}_n,\mathbf{\hat{x}}_{n-1}), \mathbf{\tilde{x}})^2,
			\label{eq:D_objective_rd}
		\end{align}
	}%
	\vspace{-0.5cm}
	{\small
		\begin{align}
			\min_\mathfrak{G}\!V_{\text{LS}}(\mathfrak{G})\!= \!&\sum_{n=1}^N\!\frac{1}{2N}\mathbb{E}_{\mathbf{z}_n\sim p_{\mathbf{z}}(\mathbf{z}),\mathbf{\tilde{x}}\sim p_{\text{data}}(\mathbf{\tilde{x}})}\!(D(G_n(\mathbf{z}_n,\mathbf{\hat{x}}_{n-1}), \mathbf{\tilde{x}})\!-\!1)^2 \nonumber\\ &+ \sum\nolimits_{n=1}^N\lambda_n|| G_n(\mathbf{z}_n,\mathbf{\hat{x}}_{n-1}) - \mathbf{x}||_1.
			\label{eq:G_objective_rd}
		\end{align}
	}%
	Unlike SEGAN, the discriminator $D$ in cases of ISEGAN and DSEGAN needs to handle imbalanced data as there are $N$ fake examples generated with respect to every real example. Therefore, it is necessary to divide the second term in (\ref{eq:D_objective_rd}) by $N$ to balance out penalization for real and fake examples misclassification. In addition, the first term in  (\ref{eq:G_objective_rd}) is also divided by $N$ to level its magnitude with that of the $\ell_1$-norm term \cite{Pascual2017}. To regulate the enhancement curriculum in multiple stages, we set $(\lambda_1, \lambda_2, \ldots, \lambda_N)$ to $(\frac{100}{2^{N-1}}, \ldots, \frac{100}{2^1}, \frac{100}{2^0})$. That is, $\lambda_n$ is set to double $\lambda_{n-1}$ while the last $\lambda_N$ is fixed to 100 as in case of SEGAN. With this curriculum, we expect the enhanced output of a generator to be twice as good as that of its preceding generator in terms of $\ell_1$-norm. As a result, the enhancement mapping learned by a generator in the chain doesn't need to be perfect as in single-stage enhancement since its output will be refined by its successor. 
	
	\vspace{-0.25cm}
	\section{Network architecture}
	\label{sec:architecture}
	\subsection{Generators $\text{G}_n$}
	\label{ssec:generator}
	\vspace{-0.05cm}
	
\begin{SCfigure}[][t]
		\centering
		\captionsetup{width=1\textwidth}
		\caption{The generator architecture used in ISEGAN and DSEGAN, featuring 11 strided convolutional layers in the encoder and 11 deconvolutional layers in the decoder.}
		\includegraphics[width=0.425\linewidth]{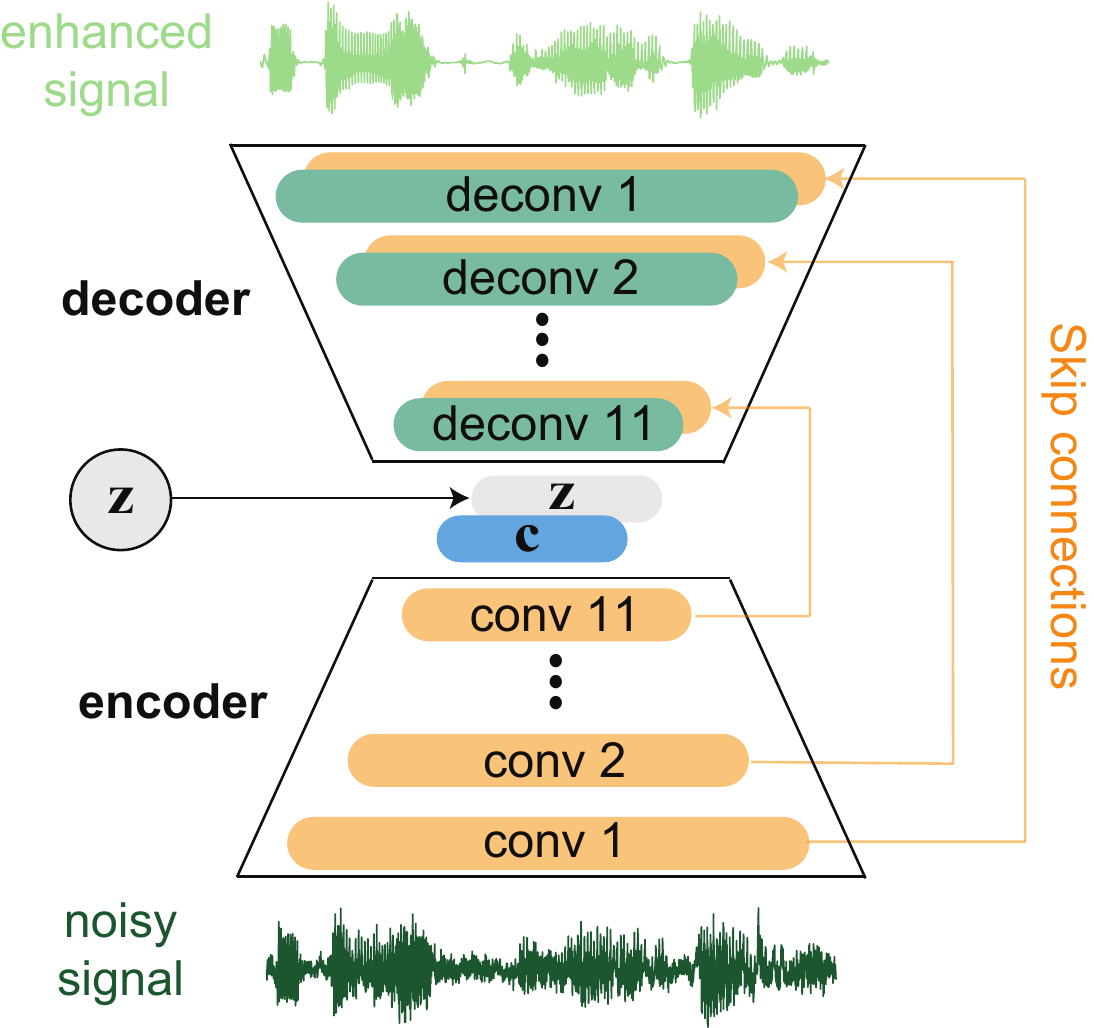}
		\vspace{-0.1cm}
		\label{fig:segan_G}
		\vspace{-0.3cm}
	\end{SCfigure}

	The architecture of the generators $G_n$, $1\!\le\!n\!\le\!N$, used in ISEGAN and DSEGAN is illustrated in Fig.~\ref{fig:segan_G}. They make use of an encoder-decoder architecture with fully-convolutional layers \cite{Radford2016}, which is similar to that used in SEGAN. Each generator receives a segment of raw signal with a length of $L\!\!=\!\!16384$ samples (approximately one second at 16 kHz) as input. The generators' encoder is composed of 11 one-dimensional strided convolutional layers with a common filter width of 31 and a stride length of 2, followed by parametric rectified linear units (PReLUs)~\cite{He2015}. The number of filters is designed to increase along the encoder's depth to compensate for the smaller and smaller convolutional output, resulting in output sizes of $8192 \times 16$, $4096 \times 32$, $2048 \times 32$, $1024 \times 64$, $512 \times 64$, $256 \times 128$, $128 \times 128$, $64 \times 256$, $32 \times 256$, $16 \times 512$, $8 \times 1024$ at the 11 convolutional layers, respectively. At the end of the encoder, the encoding vector $\mathbf{c} \in \mathbb{R}^{8 \times 1024}$ is concatenated with the noise sample $\mathbf{z} \in \mathbb{R}^{8 \times 1024}$ sampled from the normal distribution $\mathcal{N}(\mathbf{0}, \mathbf{I})$ and presented to the decoder. The generator's decoder mirrors the encoder architecture with the same number of filters and filter width (see Fig. \ref{fig:segan_G}) to reverse the encoding process by means of deconvolutions (i.e. fractional-strided transposed convolution). Note that each deconvolutional layer is again followed by a PReLU. The skip connections are employed to connect an encoding layer to its corresponding decoding layer to allow the information of the waveform to flow into the decoding stage \cite{Pascual2017}.
	
	\sidecaptionvpos{figure}{c}
	
	\vspace{-0.3cm}
	\subsection{Discriminator D}
	\vspace{-0.05cm}
	The discriminator $D$ has similar architecture to the encoder part of the generators described in  Section \ref{ssec:generator}, except that it has two-channel input and uses virtual batch-norm \cite{Salimans2016} before LeakyReLU activation with $\alpha=0.3$. In addition, $D$ is topped up with a one-dimensional convolutional layer with one filter of width one (i.e. $1\times 1$ convolution) to reduce the last convolutional output size from $8 \times 1024$ to $8$ features before classification takes place with a softmax layer.
	
	\vspace{-0.2cm}
	\section{Experiments}
	\label{sec:typestyle}
	
	\setlength\tabcolsep{1.25pt}
	\begin{table*}[t!]
		\caption{Results obtained by the studied speech enhancement systems  on the objective evaluation metrics.}
		\vspace{-0.35cm}
		\scriptsize
		\begin{center}
			\begin{tabular}{|>{\arraybackslash}m{0.4in}|>{\centering\arraybackslash}m{0.35in}|>{\centering\arraybackslash}m{0.35in}|>{\centering\arraybackslash}m{0.4in}||>{\centering\arraybackslash}m{0.65in}|>{\centering\arraybackslash}m{0.725in}|>{\centering\arraybackslash}m{0.725in}|>{\centering\arraybackslash}m{0.725in}|>{\centering\arraybackslash}m{0.725in}|>{\centering\arraybackslash}m{0.725in}|>{\centering\arraybackslash}m{0.725in}|}
				\hline
				\multirow{2}{*}{Metric}  & \multirow{2}{*}{Noisy} & \multirow{2}{*}{DNN \cite{Xu2015}} & \multirow{2}{*}{TSN \cite{Kim2019}}  & \multirow{2}{*}{SEGAN} & \multicolumn{3}{c|}{ISEGAN} & \multicolumn{3}{c|}{DSEGAN} \parbox{0pt}{\rule{0.pt}{0ex+\baselineskip}}\\ [0ex] 	
				\cline{6-11}
				&  & & &  & $N=2$ & $N=3$ & $N=4$ & $N=2$ & $N=3$ & $N=4$ \parbox{0pt}{\rule{0.pt}{0ex+\baselineskip}}\\ [0ex] 	
				\hline
				PESQ & $1.97$ & $2.45$ & $2.68$ & $2.19 \pm 0.04$ & $\bm{2.24 \pm 0.05}$ & $2.19 \pm 0.04$ & $\bm{2.21 \pm 0.06}$ & $\bm{2.35 \pm 0.06}$ & $\bm{2.39 \pm 0.02}$ & $\bm{2.37 \pm 0.05}$ \parbox{0.5pt}{\rule{0pt}{0ex+\baselineskip}}\\ [0ex] 	
				CSIG & $3.35$ & $3.73$ & $3.96$ & $3.39 \pm 0.03$ & $3.23 \pm 0.10$ & $2.96 \pm 0.08$ &  $3.00 \pm 0.14$ & $\bm{3.55 \pm 0.06}$ & $\bm{3.46 \pm 0.05}$ & $\bm{3.50 \pm 0.01}$ \parbox{0.5pt}{\rule{0pt}{0ex+\baselineskip}}\\ [0ex] 	
				CBAK & $2.44$ & $2.89$ & $2.94$ & $2.90 \pm 0.07$ & $\bm{2.95 \pm 0.07}$ & $2.88 \pm 0.12$ & $\bm{2.92 \pm 0.06}$ & $\bm{3.10 \pm 0.02}$ & $\bm{3.11 \pm 0.05}$ & $\bm{3.10 \pm 0.04}$ \parbox{0.5pt}{\rule{0pt}{0ex+\baselineskip}}\\ [0ex] 	
				COVL & $2.63$ & $3.09$ & $3.32$ & $2.76 \pm 0.03$ & $2.69 \pm 0.05$ & $2.52 \pm 0.04$ & $2.55 \pm 0.09$ & $\bm{2.93 \pm 0.05}$ & $\bm{2.90 \pm 0.03}$ & $\bm{2.92 \pm 0.02}$ \parbox{0.5pt}{\rule{0pt}{0ex+\baselineskip}}\\ [0ex] 	
				SSNR & $1.68$ & $3.64$ & $2.89$ & $7.36 \pm 0.72$ & $\bm{8.17 \pm 0.69}$ & $\bm{8.11 \pm 1.43}$ & $\bm{8.86 \pm 0.42}$ & $\bm{8.70 \pm 0.34}$ & $\bm{8.72 \pm 0.64}$ &  $\bm{8.59 \pm 0.49}$ \parbox{0.5pt}{\rule{0pt}{0ex+\baselineskip}}\\ [0ex] 	
				STOI & $92.10$ & $89.14$ & $92.52$ & $93.12 \pm 0.17$ & $\bm{93.29 \pm 0.16}$ & $\bm{93.35 \pm 0.08}$ & $\bm{93.29 \pm 0.19}$ & $\bm{93.25 \pm 0.17}$ & $\bm{93.28 \pm 0.17}$ & $\bm{93.49 \pm 0.09}$ \parbox{0.5pt}{\rule{0pt}{0ex+\baselineskip}}\\ [0ex] 	
				\hline
			\end{tabular}
		\end{center}
		\label{tab:layers}
		\vspace{-0.6cm}
	\end{table*}
	
	\vspace{-0.25cm}
	\subsection{Dataset}
	\label{ssec:dataset}
	\vspace{-0.05cm}
	To assess the performance of the proposed ISEGAN and DSEGAN and demonstrate their advantages over SEGAN, we conducted experiments on the database in \cite{Botinhao2016} which was used to evaluate SEGAN in \cite{Pascual2017}. The database is originated from the Voice Bank corpus \cite{Veaux2013} and consists of data from 30 speakers. Following the database's original split, data from 28 speakers was used for training and data from two remaining speakers was used for testing.
	
	A total of 40 noisy conditions was made in the training data by combining ten types of noises (two artificial and eight stemmed from the Demand database \cite{Thiemann2013}) with four signal-to-noise ratios (SNRs) each: 15, 10, 5, and 0 dB. For the test data, 20 noisy conditions were created, combining five types of noise from the Demand database with four SNRs each: 17.5, 12.5, 7.5, and 2.5 dB. There are about 10 and 20 utterances for each noisy condition per speaker in the training and test set, respectively. All utterances were downsampled to 16 kHz.
	
	\vspace{-0.3cm}
	\subsection{Baseline system}
	\vspace{-0.05cm}
	SEGAN was used as a baseline for comparison. We repeated training SEGAN to ensure a similar experimental setting across systems. In addition, to shed some light on how generative models like ISEGAN and DSEGAN perform on the speech enhancement task in relation to discriminative models, we also compared the proposed method to two discriminative deep learning methods: (1) the popular DNN proposed in \cite{Xu2015} and (2) the two-stage network (TSN) recently proposed in \cite{Kim2019}. 
	The DNN baseline was implemented based on \cite{Xu2015}, but with three main modifications: (a) wideband operation (16 kHz, leading to doubling of the feature dimension), (b) smaller frame size and shift (25 ms and 10 ms, respectively), and (c) use of the Adam optimizer \cite{Kingma2015} and simplified training (i.e. without unsupervised pre-training). In addition, early stopping was carried out during training via a leave-out validation set (10\% of the training data). While these modifications may lead to a better baseline, they also allow a fair comparison with the SEGAN-based systems. The TSN baseline was configured based on \cite{Kim2019}, except for the use of wideband speech. For both the baselines, the features (log-power spectra) were normalized at utterance level to zero mean and unit standard deviation. De-normalization was then performed before waveform reconstruction. The utterance-based mean and standard deviation computed from the input noisy features were used for both normalization and de-normalization.
	
	\vspace{-0.3cm}
	\subsection{Network parameters}
	\label{ssec:dataset}
	\vspace{-0.05cm}
	
	The implementation was based on Tensorflow framework \cite{Abadi2016}. The networks were trained for 100 epochs with RMSprop optimizer \cite{Tieleman2012} and a learning rate of $0.0002$. The SEGAN baseline was trained with a minibatch size of 100 while it was reduced to 50 to train ISEGAN and DSEGAN to cope with their larger memory footprints.  We experimented with different values for $N=\{2, 3, 4\}$ to investigate the influence of the number of iterations of ISEGAN and the depth of DSEGAN.
	
	As in \cite{Pascual2017}, during training, raw speech segments of length 16384 samples were extracted from the training utterances with 50\% overlap. A high-frequency preemphasis filter of coefficient $0.95$ was applied to each signal segment before presenting to the networks. During testing, raw speech segments were extracted from a test utterance without overlap. They were processed by a trained network, deemphasized, and eventually concatenated to produce the enhanced utterance. 
	\vspace{-0.25cm}
	\subsection{Objective evaluation}
	\label{ssec:objective_metrics}
	\vspace{-0.05cm}
	
	We quantified the quality of the enhanced signals based on five objective signal-quality metrics, including PESQ, CSIG, CBAK, COVL, and SSNR, as suggested in \cite{Loizou2013} and the speech-intelligibility measure STOI \cite{Taal2011}. The tool used for computing the first five metrics is based on the implementation in \cite{Loizou2013}. This is also the one used in \cite{Pascual2017}. The metrics were computed for each system by averaging over all 824 files of the test set. Since we found that the performance may vary with different network checkpoints, the mean and standard deviation of each metric over the 5 latest network checkpoints are reported.
	
	The objective evaluation results are shown in Table \ref{tab:layers}. As expected, SEGAN enhances the noisy signals to result in speech signals with better quality and intelligibility, evidenced by its better results across the objective metrics compared to those measured from the noisy signals. In comparision to SEGAN, on the one hand, ISEGAN performs comparably in terms of speech-quality metrics, slightly surpassing the baseline in PESQ, CBAK, and SSNR (i.e. with $N\!=\!2$ and $N\!=\!4$) but marginally underperforming in CSIG and COVL. On the other hand, DSEGAN obtains the best results, consistently outperforming both SEGAN and ISEGAN across all the speech quality metrics. For example, with $N=2$, DSEGAN leads to relative improvements of $7.3\%$, $4.7\%$, $6.9\%$, $6.2\%$, and $18.2\%$ over the baseline on PESQ, CSIG, CBAK, COVL, and SSNR, respectively. In terms of speech intelligibility, ISEGAN and DSEGAN obtain similar STOI results and both of them outperform SEGAN on this metric. The results in the table also suggest marginal impact of ISEGAN's number of iterations and DSEGAN's depth larger than $N\!=\!2$ since no significant performance improvements are seen.
	
	\begin{figure} [!t]
		\centering
		\includegraphics[width=0.85\linewidth]{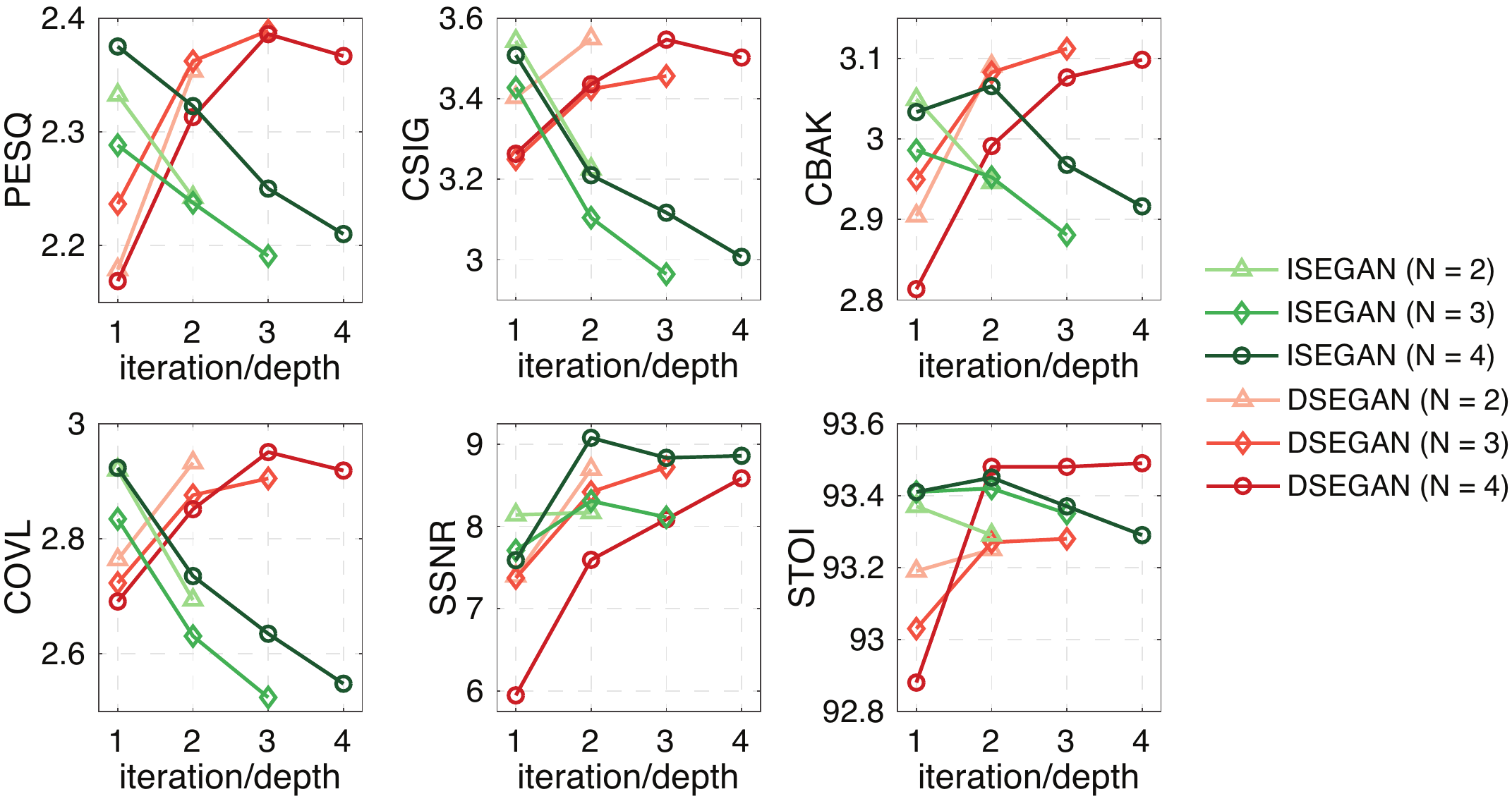}
		\vspace{-0.15cm}
		\caption{Evolution of the evaluation metrics along the depth and iteration of DSEGAN and ISEGAN, respectively.}
		\label{fig:metrics_evolution}
		\vspace{-0.3cm}
	\end{figure}
	
	Interestingly, quite opposite results are seen between the discriminative baselines (DNN and TSN) and the generative models (ISEGAN and DSEGAN). In terms of speech quality, the discriminative models outperform the generative counterparts on PESQ, CSIG, COVL but underperform on CBAK and especially on SSNR. In addition, both DNN and TSN perform poorly on speech intelligibility. Degradation on STOI metric is even seen by DNN while TSN brings up modest improvement. On the contrary, both ISEGAN and DSEGAN obtain far better results on speech intelligibility. These results suggest that the discriminative models may alter the noisy input more aggressively than the generative ones and, as a result, introduce more artifacts to the enhanced signals.
	
	To shed light on how the perfomance evolves during the enhancement process of DSEGAN and ISEGAN, we extracted and evaluated the output signals after each of their generators. The results are shown in Fig. \ref{fig:metrics_evolution}. One can observe diverging patterns between DSEGAN and ISEGAN. With DSEGAN, overall, the enhancement performance is gradually improved when the signal is passed though the generators one after another. On the contrary, ISEGAN exposes a downward trend on most of the metrics with further enhancement iterations, except for SSNR. The rationale behind the SSNR improvement is that this measure best reflects the least-squares loss that was used to train the network. However, the improved SSNR does not properly reflect other metrics such as human perception and intelligibility represented by PESQ and STOI, which rely on frame-wise weighted frequency domain. This result tends to agree with the finding in psychoacoustics \cite{Thiede2000}. We speculate that parameter independency/sharing is the key. With independent parameters, each DSEGAN's generators is tasked for enhacement with one condition of noise and has full freedom to adapt to it. On the other hand, parameter sharing forces the common generator of ISEGAN to deal with all conditions of noise, which is hard to achieve. Of note, instead of using all generators as a whole (i.e. the results in Table \ref{tab:layers}), output of any generators can be used for inferencing. For ISEGAN, using the outputs of earlier generators for this purpose is apparently reasonable as suggested in Fig. \ref{fig:metrics_evolution}.
	\vspace{-0.3cm}
	\subsection{Subjective evaluation}
	
	To validate the objective evaluation, we conducted a small-scale subjective evaluation of four conditions: noisy signals, SEGAN, ISEGAN and DSEGAN signals (with $N\!=\!2$). Twenty volunteers aged 18--52 (F=6, M=14), with self-reported normal hearing, were asked to provide forced binary quality assessments between pairs of 20 randomly presented sentences, balanced in terms of speakers and noise types, i.e. each comparison varied only in the type of system. Following a familiarization session, tests were run individually using MATLAB, with listeners wearing Philips SHM1900 headphones in a low-noise environment.
	For each pair of utterances, the selected higher quality one was rewarded $1.0$ while the lower quality received no reward. A \emph{preference} score was obtained for each system by dividing its accumulated reward by the count of its occurrences in the test. Due to the small sample size, we assessed statistical significance of results using $t$-test.
	Results confirm that the three SEGAN signals are perceived as higher quality than the noisy signals ($0.55$ to $0.45$, with $p<\!0.05$). DSEGAN and ISEGAN together significantly outperform SEGAN ($0.67$ to $0.33$, $p\!<\!0.001$). However, DSEGAN and ISEGAN qualities were not significantly different ($0.48$ to $0.52$) in this small test. Results support the detailed objective evaluation in which DSEGAN performs much better than either SEGAN or noise, however we find that ISEGAN also performs well in subjective tests.
	
	
	\vspace{-0.3cm}
	\section{Conclusions}
	\label{sec:conclusions}
	\vspace{-0.1cm}
	
	This paper presented a GAN method with multiple generators to tackle speech enhancement. Using multiple chained generators, the method aims to learn multiple enhancement mappings, each corresponding to a generator in the chain, to accomplish a multi-stage enhancement process. Two new architectures, ISEGAN and DSEGAN, were proposed. ISEGAN's generators share their parameters and, as a result, are constrained to learn a common mapping for all the enhancement stages. DSEGAN, in contrast, has independent generators that allow them to learn different mappings at different stages.  Objective tests demonstrated that the proposed ISEGAN and DSEGAN perform comparably and are better than SEGAN on speech-quality metrics and that learning independent mappings leads to better performance than a common mapping. In addition, both the proposed systems achieve more favourable results than SEGAN on the speech-intelligibility metric as well as the subjective perceptual test.
	
	\bibliographystyle{IEEEbib}
	\bibliography{refgan}

\end{document}